# A Multi-module Robust Method for Transient Stability Assessment against False Label Injection Cyberattacks

Hanxuan Wang[1,3], Na Lu[1]*, Yinhong Liu[1], Zhuqing Wang[2,3], Zixuan Wang[1]

1 Systems Engineering Institute, School of Automation Science and Engineering, Xi'an Jiaotong University, Xi'an, China 710049

2 School of Computer Science, Northwestern Polytechnical University, Xi'an, China 710129

3 College of Computing and Data Science, Nanyang Technological University, Singapore, 639798

## Abstract

The success of deep learning in transient stability assessment (TSA) heavily relies on high-quality training data. However, the label information in TSA datasets is vulnerable to contamination through false label injection (FLI) cyberattacks, resulting in degraded performance of deep TSA models. To address this challenge, a <u>M</u>ulti-<u>M</u>odule <u>R</u>obust TSA method (MMR) is proposed to rectify the supervised training process misguided by FLI in an unsupervised manner. In MMR, a supervised classification module and an unsupervised clustering module are alternatively trained to improve the clustering friendliness of representation leaning, thereby achieving accurate clustering assignments. Leveraging the clustering assignments, we construct a training label corrector to rectify the injected false labels and progressively enhance robustness and resilience against FLI. However, there is still a gap on accuracy and convergence speed between MMR and FLI-free deep TSA models. To narrow this gap, we further propose a human-in-the-loop training strategy, named MMR-HIL. In MMR-HIL, potential false samples can be detected by modeling the training loss with a Gaussian distribution. From these samples, the most likely false samples and most ambiguous samples are re-labeled by a TSA experts guided bi-directional annotator and then subjected to penalized optimization, aimed at improving accuracy and convergence speed. Extensive experiments indicate that MMR and MMR-HIL both exhibit powerful robustness against FLI in TSA performance. Moreover, the contaminated labels can also be effectively corrected, demonstrating superior resilience of the proposed methods.

| **Nomenclature** | | | |
|---|---|---|---|
| ***Problem formulation*** | | | |
| $D$ | Clean TSA training dataset | $N$ | Training dataset sample size |
| $X$ | Transient response trajectories | $Y$ | Training labels |
| $d$ | Transient response trajectory dimension | $C$ | Number of classes |
| $D_F$ | Contaminated TSA training dataset | $\tilde{Y}$ | Contaminated training labels |
| $G$ | Noise transition matrix | $\upsilon$ | Injection ratio |
| $\eta$ | Learning rate | | |
| ***The proposed method*** | | | |
| $f_{Enc}$ | Encoder | $f_{Dec}$ | Decoder |
| $f_C$ | Classifier | $f_{Clu}$ | Clustering layer |
| $L_{Rec}$ | Reconstruction loss | $Z_e$ | Embedding space dimension |
| $L_C$ | Classification loss | $S_{intra}$ | Intra-class separation degree |
| $\alpha_1$ | Balance coefficient between $L_C$ and $L_{Rec}$ | $S_{inter}$ | Inter-class separation degree |
| $L_{CM}$ | Classification module objective function | $\bar{\mu}$ | Average embedding vector |
| $\mu$ | Clustering center vector | $m$ | Fuzzifier |
| $\alpha_2$ | Balance coefficient between $L_{Clu}$ and $L_{Rec}$ | $q$ | Soft clustering assignments |
| $L_{Clu}$ | Clustering loss | $p_t$ | Target distribution |
| $L_{CluM}$ | Clustering module objective function | $Y_C$ | Classification predictions |
| $\kappa$ | Correction coefficient | $Y_{Clu}$ | Hard clustering assignments |
| $p_{false}$ | Probability of being a false sample predicted by GMM | $\tau$ | False label detection threshold |
| | | $\rho$ | Annotation rate |
| $\varepsilon$ | Penalized coefficient | $T$ | Annotation frequency |
| ***Case studies*** | | | |
| $\xi$ | Relative efficiency | $\xi^*$ | Absolute efficiency |
| $k$ | Convergence epoch increment | $\Delta$ | Accuracy increment |
| $N_q$ | Proportion of query samples | $\rho_\uparrow$ | Annotation rate in ascending order |
| $N_{dq}/N_q$ | Proportion of duplicate query samples | $\rho_\downarrow$ | Annotation rate in descending order |

# 1 Introduction

Transient stability assessment (TSA) refers to the evaluation of the stability of a power system after a set of contingencies, which plays an important role in power system operation, planning and control [1, 2]. However, due to the increased penetration of renewable energy sources in modern power systems, traditional TSA methods, such as time domain simulation and transient energy function methods [2, 3], have gradually become inadequate and cannot effectively support the practical applications.

Fortunately, with the rapid development of deep learning these years [4, 5], the powerful nonlinear representation learning ability of neural networks has gained favor among researchers in the community of TSA and achieved great successes [6-8]. As reported in [9, 10], the superior performance of deep neural networks heavily relies on high-quality training data. Once the training data is contaminated, neural networks are prone to severe overfitting. This dependence makes high-quality TSA datasets a prerequisite for accurately assessing transient stability with deep models.

Most TSA datasets are sourced from power system historical operation data and computer simulations. After collection, these datasets are typically stored on servers for subsequent analysis or utilization. However, as one of the most important cyber-physical infrastructures in both industry and daily life, power system has emerged as primary targets of cyberattacks [11]. Among these attacks, the primary method of contaminating TSA datasets is through false data injection attack (FDIA). FDIA can clandestinely manipulate transient response trajectories and label information in TSA datasets stored on servers through techniques such as SQL injection [12] and Man-in-the-Middle attacks [13], resulting in significant degradation on the performance of deep TSA models. According to the type of injected information, FDIA can be categorized into false feature injection (FFI) and false label injection (FLI). In the case of FFI, the feature information (e.g., PMU measured transient response trajectories) is contaminated by injecting adversarial perturbations. A plethora of researches have emerged based on this type of attack [14-18]. Unlike FFI, in FLI, the transient stability of some samples in TSA datasets is tampered with, leading to poor generalization performance. Unfortunately, there is a relatively limited research on FLI in the area of TSA [19]. Indeed, due to the destructive impact on neural network performance, FLI has garnered the attention of machine learning researchers. Specifically, the robust learning methods against FLI can be generally grouped into four categories: (1) robust architecture-based approach, (2) robust regularization approach-based approach, (3) sample selection-based approach, and (4) label correction-based approach [10]. The robust architecture-based approaches refer to developing reliable architectures to learn the label transition process and correcting false labels according to the estimated noise transition properties [20-22]. The performance of these methods relies on the accurate estimation of the noise transition matrix, which requires prior information about FLI, limiting the potential of robust architecture-based approaches. The robust regularization-based approaches aim to explicitly or implicitly regulate the training process of a network to mitigate the overfitting issue brought by FLI [23-25]. While these methods can provide theoretical guarantees, they suffer from underfitting, resulting in decreased performance of TSA models. For sample selection, these methods identify true-labeled samples from contaminated training data via multi-network or multi-round learning [19, 26-28]. These methods have become one of the most popular solutions to FLI due to their superior performance. However, after detecting false samples, whether by training only with clean samples or transitioning to semi-supervised learning, the model performance is limited due to the abandonment of label information. Label correction-based approaches, as a new technique in robust learning, directly rectify potential false labels using the geometric and probabilistic properties of the training samples [29, 30]. Correcting false labels enables these methods to fundamentally improve the robustness of deep TSA models against FLI and our proposed MMR also falls into this category. In summary, despite the FLI problem has received some attention in the field of machine learning, it still lacks sufficient focus in the community of TSA. Identifying the correct latent TSA patterns from training data contaminated by FLI still remains an unsolved challenge.

To address the false label injection problem in TSA, a multi-module robust transient stability assessment method MMR is proposed. Its main idea is to rectify the misguided supervised training caused by FLI through unsupervised clustering methods which is trained independently from the contaminated training labels. Specifically, there are three modules in MMR: classification module, clustering module and training label corrector. The classification module possesses strong representation learning ability; however, it is easily influenced by injected false labels. In contrast, the clustering module can predict clustering assignments, i.e., transient stability, independently of label information but suffers from weak representation learning ability, resulting in poor accuracy of the clustering assignments. We alternatively train the classification module and the clustering module to improve the clustering friendliness of representation leaning, thereby achieving accurate clustering assignments. By integrating the classification predictions with the clustering assignments, the injected false labels can be rectified and the robustness and resilience of deep TSA models against FLI can be improved gradually. Furthermore, to improve convergence speed and assessment accuracy to the level of FLI-free deep TSA models, a human-in-the-loop training strategy MMR-HIL is introduced. In MMR-HIL, the classification loss can be modeled as the probability of being false samples using Gaussian distributions. With the introduction of a bi-directional annotator, we can identify the most likely false samples and the most ambiguous samples based on the modeled probability. These identified samples can then be re-labeled through querying TSA experts, followed by training with penalized reweight optimization. As a result, the convergence speed and accuracy can be effectively improved. It is noteworthy that while all the neural network modules and objective functions used in MMR and MMR-HIL are popular techniques, the novelty of this work lies in the introduction of a novel robust learning framework. This framework aims to alleviate the adverse effect of FLI in TSA by employing unsupervised learning to correct the misled supervised learning. The contributions of this paper are summarized as follows:

(1) A multi-module robust transient stability assessment method MMR is proposed, which rectifies the FLI misguided supervised classification training process with unsupervised clustering methods in an alternating training way.

(2) We develop a training label corrector to rectify the false labels in the training data by integrating the supervised classification predictions with the unsupervised clustering assignments, which can improve the robustness against FLI fundamentally.

(3) A human-in-the-loop training strategy, MMR-HIL, is designed to further improve the convergence speed and assessment accuracy of MMR to the level of FLI-free scenarios.

The remainder of this paper is organized as follows: Section 2 introduces the problem formulation of false label injection in TSA. Section 3 describes the proposed MMR and MMR-HIL. Case studies are presented in Section 4 to verify the effectiveness of the proposed methods. Finally, Section 5 concludes with a discussion.

## 2 Problem Formulation

Transient stability assessment can be modeled as a classification task for learning a specific function which maps the transient response trajectories (input features) to the corresponding transient stability (labels). Taking power angle stability [6, 31, 32] as the object of this research, we consider a clean TSA dataset $D = \{(x_i, y_i)\}_{i=1}^{N}$, where $N$ is the sample size of this dataset, $X$ is the feature space of the transient response trajectories and $Y$ is the ground-truth label space of the transient stability. The primary objective of TSA is to determine a mapping function $f(\cdot;\Theta): X \to Y$ parameterized by $\Theta$, where the learned parameters $\Theta$ can minimize the empirical risk

$$\mathcal{R}_D(f) = \mathbb{E}_D\left[\ell(f(x;\Theta), y)\right] \approx \frac{1}{|D|} \sum_{(x,y) \in D} \ell(f(x;\Theta), y), \tag{1}$$

where $\ell$ is a classification loss function for TSA, such as cross-entropy. However, due to the occurrence of FLI, the injected false labels $\tilde{Y}$ contaminate the clean training labels in $D$, resulting in $D_F = \{(x_i, \tilde{y}_i)\}_{i=1}^{N}$. If we continue to follow the standard training process of TSA classification, a mini-batch $B$ can be sampled randomly from $D_F$ to update the parameters of the deep TSA model via

$$\Theta \leftarrow \Theta - \eta \nabla \left( \frac{1}{|B|} \sum_{(x,\tilde{y}) \in B} \ell(f(x;\Theta), \tilde{y}) \right), \tag{2}$$

where $\eta$ is the learning rate. If false labels are present in this batch, the gradient will be misguided in a wrong direction, potentially leading the deep TSA model to memorize the false labels. Consequently, samples with false labels will be pushed towards regions with mismatched stability in the embedding space. Such distortion within the embedding space can result in overly complex decision boundaries, ultimately degrading the generalization ability.

Fig. 1 illustrates how false labels mislead gradients and degrade the generalization ability, where the shapes (circle and triangle) symbolize the true transient stability and the colors (red and blue) correspond to the contaminated labels. Fig. 1(a) depicts the misled optimization gradients caused by the injected false labels. For instance, considering the sample enclosed by the black rectangle, its false label compels it to follow the incorrect gradient direction $\nabla_F$, whereas it should have been optimized along the correct gradient direction $\nabla_C$. These misled gradients contribute to the distortion in the embedding space. Fig. 1(b) shows the degradation in generalization ability caused by FLI. The black triangles and circles denote testing samples, while the purple and yellow curves represent the decision boundaries corresponding to clean and contaminated labels, respectively. FLI can result in overly complex decision boundaries, leading to severe overfitting, which weakens the model's ability to accurately assess the transient stability.

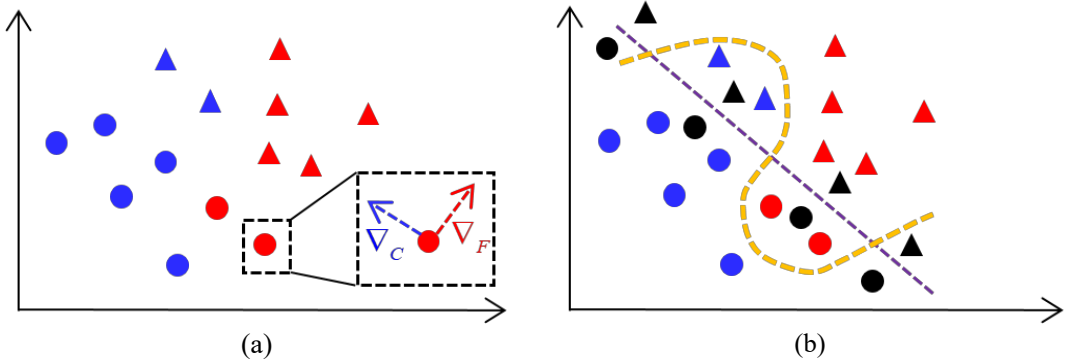

**Fig. 1** The illustration of how FLI degrades the generalization ability of deep TSA models. (a) depicts the misled optimization direction of the samples with false labels. (b) illustrates the overfitted decision boundaries.

According to the distribution of the injected false labels [10], FLI can be categorized into symmetric and asymmetric (Sym/Asym-FLI). Assuming the true labels are corrupted by a noise transition matrix $G \in [0,1]^{C \times C}$, where $G_{ij} = p(\tilde{y}=j|y=i)$ represents the probability of a sample with true label $i$ being corrupted into the $j$-th class. For Sym-FLI, the true labels are flipped to

any of the other labels with probability $\upsilon \in [0,1]$ and $\forall_{i=j} G_{ij} = 1-\upsilon \wedge \forall_{i \neq j} G_{ij} = (\upsilon/(C-1))$. In this research, given that TSA is a binary classification problem, Sym-FLI flips the labels of stable and unstable samples at an equal injection ratio. In the case of Asym-FLI, the corrupted labels are flipped between semantically similar classes, i.e., $\forall_{i=j} G_{ij} = 1-\upsilon \wedge \exists_{i \neq j} G_{ij} = \upsilon$. Since misclassifying unstable samples in TSA as stable may pose a serious threat to the security of power systems, Asym-FLI randomly flips the labels of unstable samples to stable ones, while keeping the stable samples unchanged.

This study aims to excavate accurate latent patterns from TSA data with contaminated labels and improve the robustness of deep TSA models against false label injection. There are several challenges in this task:

(1) Imperceptible: The ground truth of the contaminated TSA data remains unaccessible during training.

(2) Confidential: The type, ratio and even the presence of FLI are unknown.

(3) Unvalidated: Creating a reliable validation set for evaluating model performance becomes challenging in the presence of false labels, which complicates the evaluation of overfitting.

## 3 The Proposed Methods

The proposed methods are organized as follows: in Section 3.1, we introduce a multi-module robust transient stability assessment method, MMR, which is designed to rectify the misguided supervised training caused by FLI in an unsupervised manner. In Section 3.2, we propose a human-in-the-loop training strategy, MMR-HIL, aiming to improve the assessment accuracy and convergence speed of MMR to the level of FLI-free scenarios. Section 3.3 presents a unified framework for MMR and MMR-HIL, along with complexity analysis. The overall schema of MMR and MMR-HIL is shown in Fig. 2.

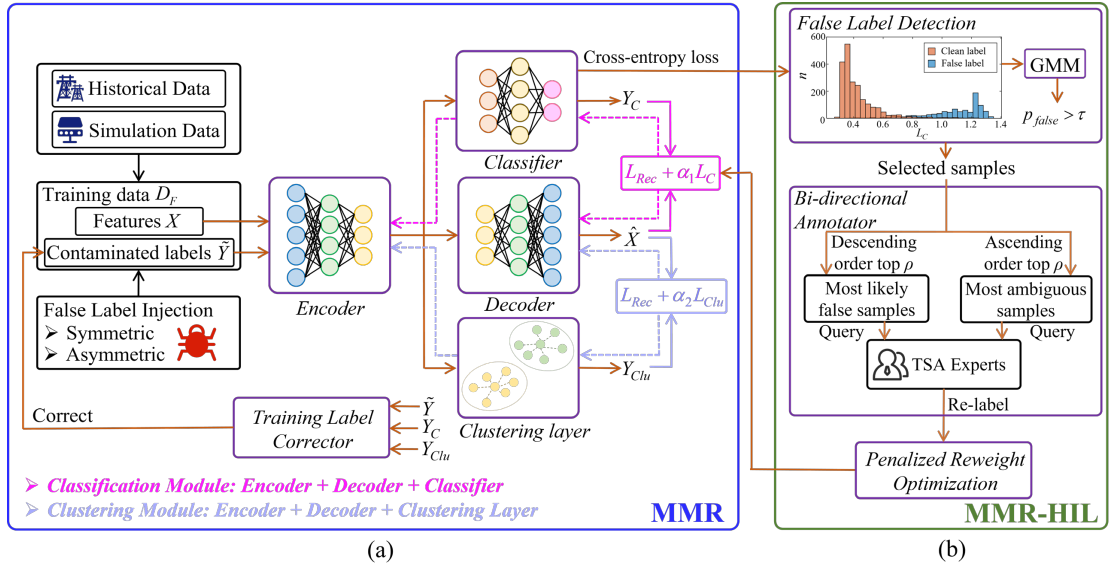

**Fig. 2** The overall schema of (a) MMR and (b) MMR-HIL. Solid lines represent forward propagation, while dashed lines denote backward propagation.

## 3.1 Multi-module Robust Transient stability Assessment

Given a TSA training dataset contaminated by FLI, our objective is to excavate the correct TSA patterns from the interference of the injected false labels. Previous studies on robust learning [28-30, 33, 34] suggest that the most direct and effective approach to improve robustness against FLI is to correct false labels during the training process. However, the imperceptible nature of FLI renders the attack information inaccessible to the training process. Fortunately, unsupervised clustering can predict clustering assignments, i.e., transient stability, without using any label information, providing a promising solution for rectifying the injected false labels. Based on the above considerations, we design MMR to combat the adverse influence of FLI on TSA. MMR is structured with three modules: classification module, clustering module and training label corrector as illustrated in Fig 2(a). During the training, the classification module and the clustering module are trained alternatively to constrain the distortion of false labels on the embedding space and enhance the clustering-friendliness of the embedding features, improving the accuracy of the solved clustering assignments. In each epoch, the training label corrector rectifies false labels by integrating the classification predictions and the clustering assignments, gradually improving robustness and resilience against FLI. The detailed descriptions of each module are as follows:

➢ *Classification Module*

Considering a training dataset with $N$ samples contaminated by FLI denoted as $D_F = \{(x_i, \tilde{y}_i)\}_{i=1}^{N}$, where $x_i$ is the $i$-th transient response trajectory and $\tilde{y}_i$ is the corresponding transient stability (might be false). The real transient stability status remains imperceptible during the training. Firstly, the transient response trajectories collected by PMUs typically have high dimensions in both the time and spatial domains. Due to the Curse of Dimensionality [35], these input features can degrade the accuracy of distance measurements in both classification and clustering tasks. Therefore, it is necessary to perform dimension reduction on the transient response trajectories. Motivated by this necessity, a convolutional autoencoder is introduced, i.e., the encoder and the decoder in Fig. 2(a). The encoder maps the high-dimensional transient response trajectories into the low-dimensional embedding features, while the decoder reconstructs these embedding features back to their original inputs. Effective reconstruction of the inputs would result in the embedding features forming a valuable set of representations. The objective function for the autoencoder can be expressed as

$$L_{Rec} = \frac{1}{N}\sum_{i=1}^{N} \left\| x_i - f_{Dec}\left(f_{Enc}(x_i)\right) \right\|_2^2, \qquad (3)$$

where $f_{Enc}$ and $f_{Dec}$ represent the mapping of the encoder and the decoder respectively. The low-dimensional embedding features can be extracted via $z = f_{Enc}(x)$, where $z \in \mathbb{R}^{Z_e}$ and $Z_e$ is the dimension of the embedding features.

With the embedding features as input, the objective function of the classifier is denoted as:

$$L_C = -\frac{1}{N}\sum_{i=i}^{N}\sum_{c=1}^{2}\tilde{y}_{i,c}\log\left(f_C\left(f_{Enc}(x_i)\right)_c\right), \quad (4)$$

where $f_C: \mathbb{R}^{Z_e} \to [0,1]^{C=2}$ represents the classifier with a softmax layer as the output, performing a mapping from the embedding space to the probability space. However, due to the existence of false labels in the TSA training dataset, directly using cross-entropy loss may result in severe overfitting. As defined by cross-entropy, the component contributing to the optimization of the $i$-th sample with training label $c$ is $-\tilde{y}_{i,c}\log\left(f_C\left(f_{Enc}(x_i)\right)_c\right)$. Because $\log(x)$ is unbounded, cross-entropy loss is also unbounded, i.e., $L_{C,i} \in (0, +\infty)$. In particular, the gradient of cross-entropy loss can be expressed as

$$\frac{\partial L_C\left(f_C\left(f_{Enc}(x)\right), y\right)}{\partial \theta} = -\frac{1}{\sigma_y(z)}\nabla_\theta \sigma_y(z), \quad (5)$$

where $\theta$ represents the parameters of the classification module, and $\sigma$ denotes the softmax function. Eq. (5) shows that cross-entropy tends to focus more on samples with lower confidence, resulting in larger gradients for these samples. As $\sigma(z)$ approaches 0, the cross-entropy loss approaches infinity. According to the Small-Loss assumption [37], when there are false labels in the training dataset, the confidence of false samples tends to be lower. As a result, false samples can introduce larger erroneous gradients, ultimately leading to severe overfitting. To alleviate the overfitting issue, we train the classifier, encoder and decoder jointly, referring to this combination as the ***classification module***. The objective function of the classification module is

$$L_{CM} = L_{Rec} + \alpha_1 L_C, \quad (6)$$

where $α_1$ is a balance coefficient between the reconstruction loss $L_{Rec}$ and the classification loss $L_C$. The reason behind this design is as follows: $L_{Rec}$ ensures that samples adhere to their inherent distribution in the embedding space, and the classification performance is improved through appropriate adjustments under this distribution constraint, alleviating the distortion of the embedding space caused by false labels.

➢ *Clustering Module*

The design of the classification module can only alleviate overfitting caused by FLI but cannot correct the misguided training process. Fortunately, because unsupervised clustering is independent of training labels, theoretically, if the embedding features of the training data are sufficiently conducive to clustering, completely accurate clustering assignments can be achieved. This implies that unsupervised clustering has the potential to correct false labels. Indeed, well-performed clustering assignments should aim to keep intra-cluster distances small while ensuring that inter-cluster distances are large enough [4, 38]. Inspired by fuzzy theory [39], the degree of inter- and intra-class separation can be expressed as

$$S_{inter} = \frac{1}{N} \sum_{i=1}^{N} \sum_{j=1}^{2} q_{i,j}^m (z_i - \bar{\mu})(z_i - \bar{\mu})^\top, \tag{7}$$

$$S_{intra} = \frac{1}{N} \sum_{i=1}^{N} \sum_{j=1}^{2} q_{i,j}^m (z_i - \mu_j)(z_i - \mu_j)^\top, \tag{8}$$

where $q_{ij}$ is the clustering assignment of the $i$-th sample to the $j$-th clustering center, $m>1$ is the fuzzifier, $\mu_j$ is the vector of the $j$-th clustering center in the embedding space, $\bar{\mu} = \sum_{i=1}^{N} z_i / N$ is the average embedding feature of all the TSA training data. Therefore, the objective function of this fuzzy clustering process is defined as

$$L_{Clu} = S_{intra} - S_{inter}$$
$$s.t. \sum_{j=1}^{2} q_{ij} = 1. \tag{9}$$

Eq. (9) can be optimized with Lagrange multiplier method [40-42]. By introducing Lagrange multiplier $\lambda$, we can obtain a new objective function $F$:

$$F = \sum_{i=1}^{N}\sum_{j=1}^{2} q_{ij}^{m}\left[(z_i - \bar{\mu})^2 + (z_i - \mu_j)^2\right] + \sum_{i=1}^{N} \lambda_i \left[-1 + \sum_{j=1}^{2} q_{ij}\right]. \tag{10}$$

Let $\partial F/\partial q_{ij} = 0$, $\partial F/\partial \mu_j = 0$ and $\partial F/\partial \lambda_i = 0$, the soft cluster assignment $q_{ij}$ can be solved via

$$q_{ij} = \frac{\left(\|z_i - \mu_j\|^2 - \|\mu_j - \bar{\mu}\|^2\right)^{-1/m-1}}{\sum_{k=1}^{2}\left(\|z_i - \mu_k\|^2 - \|\mu_k - \bar{\mu}\|^2\right)^{-1/m-1}}, \tag{11}$$

and $\mu_j$ can be obtained via

$$\mu_j = \sum_{i=1}^{N} q_{ij}^{m} z_i \bigg/ \sum_{i=1}^{N} q_{ij}^{m}. \tag{12}$$

Eqns. (11) and (12) are iteratively executed until Eq. (9) converges and the optimized clustering centers can be solved. Subsequently, a clustering layer can be built with the embedding features as input and the clustering assignments as output. The forward propagation of the clustering layer is Eq. (11), using the optimized clustering centers as the initialized parameters. To further improve the compactness of similar samples, a target distribution is introduced

$$p_{t,ij} = \frac{q_{ij}^2 / \left(\sum_i q_{ij}\right)}{\sum_{k=1}^{2} q_{ik}^2 / \left(\sum_i q_{ik}\right)}. \tag{13}$$

This target distribution encourages samples to move closer to the clustering center where they most likely belong [43]. After that, the encoder and the clustering layer can be optimized by narrowing the gap between the soft clustering assignments and the target distribution with KL divergence:

$$L_{Clu} = \frac{1}{N}\sum_{i=1}^{N}\sum_{j=1}^{2} q_{ij} \log \frac{q_{ij}}{p_{t,ij}}. \tag{14}$$

However, as reported in [4, 35], solely optimizing $L_{Clu}$ may lead to overfitting. In other words, excessive optimization in the probability space weakens the correlation between the low-dimensional embedding features and the training data. Therefore, similar to the classification module, reconstruction loss is employed to constrain the distribution of the embedding features. Together, the encoder, decoder, and clustering layer constitute the clustering module, the objective function of which can be formulated as

$$L_{CluM} = L_{Rec} + \alpha_2 L_{Clu}, \tag{15}$$

where $\alpha_2$ is a balance coefficient between the reconstruction loss $L_{Rec}$ and the clustering loss $L_{Clu}$. Optimizing the clustering module can improve the clustering-friendliness of the embedding features, compelling samples from the same class to cluster together. However, despite the clustering module can avoid the interference of FLI, its unsupervised nature hinders the full utilization of the representation learning capabilities of neural networks. In other words, if we only use the clustering module in MMR, the accuracy of the solved clustering assignments is low, as shown in Section 4.2.2. The classification module benefits from the strong representation learning capability but is vulnerable to false labels. Conversely, while the clustering module is independent of false labels, it suffers from poor representation learning capability. To fully leverage the advantages of both, the classification module and the clustering module are trained alternately. Through this training strategy, the clustering-friendliness of the embedding features can be significantly improved, leading to more accurate clustering assignments.

➢ *Training Label Corrector*

The alternating training strategy between the classification module and the clustering module enables us to obtain more accurate clustering assignments. Based on this, a training label corrector is designed to rectify false labels in the TSA training data to improve robustness and resilience against FLI. Specifically, during each training epoch, the classification predictions $Y_C$ and the hard clustering assignments $Y_{Clu}$ can be solved via

$$Y_C = \mathrm{argmax}\, f_C\left(f_{Enc}(X)\right), \tag{16}$$

$$Y_{Clu} = \mathrm{argmax}\, f_{Clu}\left(f_{Enc}(X)\right). \tag{17}$$

The training labels can be updated by integrating the clustering assignments and the classification predictions:

$$y_i = (1-\omega)y_i + \omega\left(y_{c,i} + y_{Clu,i}\right)/2, \tag{18}$$

where $0 \leq \omega \leq 1$ is a weighting factor for controlling the strength of label correction. $\omega$ is a function of the training epoch $t$:

$$\omega = \begin{cases} \kappa t, & \text{if } \kappa t \leq 1 \\ 1, & \text{if } \kappa t > 1 \end{cases}, \tag{19}$$

where $\kappa > 0$ is called correction coefficient. The reason for this design is during the early stages of training, the clustering-friendliness of the embedding features is low, resulting in poor clustering assignments. Therefore, it is not feasible to correct false labels at this point. However, along with the training, the alternating training between the supervised learning and unsupervised learning gradually improves the compactness of similar samples in the embedding space, improving the accuracy of the clustering assignments. At this stage, integrating the clustering assignments with the classification predictions allows us to correct the injected false labels safely. These corrected labels can further enhance the clustering-friendliness of the embedding features, forming a positive feedback loop.

The alternating training of the classification module and the clustering module, combined with the correction of false labels, gradually improve the quality of the TSA training dataset and enhances robustness and resilience against FLI. Once the training converges, the transient stability of the power system can be predicted with the classifier via Eq. (16).

**3.2 Improve MMR with a human-in-the-loop training strategy**

Despite MMR can alleviate the adverse influence of FLI, our experiments in Section 4.3.1 revealed that there is still a gap in accuracy and training speed compared to FLI-free deep TSA models. This gap is like a "Sword of Damocles", posing a threat to the stable operation of power systems. The reasons behind the gap are as follows:

(1) *On assessment accuracy*: The degradation in assessment accuracy arises from those ambiguous samples which may locate near the decision boundary. These ambiguous samples can result in inaccurate clustering assignments, hinder the correction of false labels, and consequently reduce assessment accuracy.

(2) *On convergence speed*: False labels can be gradually corrected by the training label corrector along with the training process. However, correcting on training labels changes the optimization direction of these samples, leading to a slowdown in convergence speed.

Therefore, if we can identify the true labels of these ambiguous and false samples early on, the accuracy and convergence speed of MMR can be effectively improved. As a result, active learning has become a promising solution [5, 46], which queries TSA experts the real transient stability to improve the model performance in a human-in-the-loop way. However, due to the high cost of interactions with human experts, we need to minimize the number of queries while maximizing performance improvement. Based on this, a human-in-the-loop training strategy

MMR-HIL is proposed, as shown in Fig. 2(b), which consists of three steps: (1) false sample detection, (2) re-labeling, (3) penalized reweight optimization.

➢ *False Sample Detection*

According to [26, 37, 47], neural networks tend to learn clean and simple patterns before overfitting false and ambiguous samples due to the memorization effect. This phenomenon is known as Small-Loss assumption. Cross-entropy loss can quantify the learning difficulty at the sample level via

$$L_{C,i} = -\frac{1}{2}\sum_{j=1}^{2} \tilde{y}_{i,j} \log f_C\left(f_{enc}(x_i)\right)_j. \tag{20}$$

The collected losses of all $N$ training samples serve as the input of a Gaussian Mixture Model [48], enabling the prediction of the probability $p_{false}$ of a sample being falsely labeled. A threshold $\tau$ is then introduced to divide the training data into false samples ($p_{false}>\tau$) and clean samples ($p_{false}<\tau$). Following the setting in previous studies [26, 47], $\tau$ is set as 0.8.

➢ *Re-labeling*

Apparently, if all the detected false data are re-labeled by TSA experts, the model can exhibit great performance on both assessment accuracy and convergence speed. However, the prohibitively expensive annotation cost makes this approach unacceptable. To achieve high cost-effectiveness regarding accuracy and convergence speed, a bi-directional annotator is proposed. We sample the top $\rho$ percent of the detected false samples in descending order and ascending order of $p_{false}$, and $\rho$ is called annotation rate. All selected samples are sent to TSA experts for relabeling. Samples selected in descending order are more likely to have false labels. Although the training label corrector may gradually rectify these false samples without intervention from TSA experts, manual labeling can accelerate the training process. Conversely, samples selected in ascending order with relatively lower $p_{false}$ are ambiguous false samples which are difficult to correct. It should be mentioned that the re-labeling is performed every $T$ epoch to enhance cost-effectiveness.

➢ *Penalized Reweight Optimization*

Given the reliability of TSA expert annotations, any classification predictions conflicting with these annotations must be heavily penalized. Specifically, a penalized reweight training method is used to modify the optimization objectives for the classification module, which can be formulated as

$$L_{CM-fixed} = L_{Rec} - \frac{\alpha_1}{N}\sum_{i=i}^{N}\varepsilon_i\sum_{c=1}^{2}\hat{y}_{i,c}\log\left(f_C\left(f_{enc}(x_i)\right)_c\right), \tag{21}$$

where $\hat{y}$ represents the label of the expert-annotated training data and the penalized coefficient $\varepsilon_i$ equals 3 when $x_i$ is annotated by experts and 1 for other samples. In this way, the convergence speed and assessment accuracy can be effectively improved with the help of TSA experts.

## 3.3 Unified Framework and Complexity Analysis

In fact, MMR-HIL is a plugin for MMR. Therefore, MMR and MMR-HIL can be integrated into a unified framework, illustrated in Algorithm 1 and Fig. 2. Besides, an analysis of the complexity of MMR and MMR-HIL is provided. Firstly, the neural networks in MMR consist of an encoder $f_{Enc}$, a decoder $f_{Dec}$, a classifier $f_C$ and a clustering layer $f_{Clu}$. Let $H$ denote the maximum number of neurons in all hidden layers of $f_{Enc}$ and $f_{Dec}$, and let $Z_e$ denote the dimension of the embedding features extracted by $f_{Enc}$. Generally, $C, Z_e \ll H, N$ holds. Both complexities of $f_{Clu}$ and the solution of the target distribution are $O(NZ_eC)$. The complexities to count $f_C$ and the combination of $f_{Enc}$ and $f_{Dec}$ are $O(NH)$ and $O(NH^2)$, respectively. Therefore, the complexity of MMR scales linearly with the sample size $N$. Compared with MMR, MMR-

HIL only requires an additional traverse of the cross-entropy loss during false sample detection, leading to a complexity of $O(NC)$. As a result, the corresponding complexity is also linear to $N$.

---
**Algorithm 1: Unified Framework of MMR and MMR-HIL.**
---
**Input:** Training dataset $D_F$, Encoder $f_{Enc}$, Decoder $f_{Dec}$, Classifier $f_C$, Clustering layer $f_{Clu}$, Balance coefficients $\alpha_1$, $\alpha_2$, Correction coefficient $\kappa$.
**Output:** Well-trained $f_{Enc}$, $f_{Dec}$, $f_C$ and $f_{Clu}$.
1: $t=0$;
2: $\varepsilon=\mathbf{1}_N$;
3: **While not convergence:**
4:   **if** $t=0$ **do:**
5:     Initialize clustering layer with Eqns. (9), (11), (12);
6:   **end if**
7:   Optimize the classification module ($f_{Enc}$, $f_{Dec}$, $f_C$) with Eq. (6);
8:   **if MMR-HIL:**
9:     **if** $t$ mod $T=0$ **do:**
10:       Detecting potential false labels with Eq. (20) and GMM;
11:       Relabel the detected false labels with the bi-directional annotator.
12:       Modify the fixing weight $\varepsilon$;
13:       Optimize the classification module ($f_{Enc}$, $f_{Dec}$, $f_C$) with Eqns. (4) and (21);
14:     **end if**
15:   **end if**
16:   Solve the target distribution with Eq. (13);
17:   Optimize the clustering module ($f_{Enc}$, $f_{Dec}$, $f_{Clu}$) with Eq. (15);
18:   Solve the weighting factor $\omega$ with Eq. (19);
19:   Correct the training labels with Eqns. (16), (17), (18) and $\omega$;
20:   $t=t+1$
21: **Return** $f_{Enc}$, $f_{Dec}$, $f_C$ and $f_{Clu}$

## 4 Case Studies

In this section, the effectiveness of the proposed MMR and MMR-HIL has been evaluated on a regional power system in China under different kinds of contingencies. Section 4.1 introduces the detailed information of the datasets and the experimental setup. Section 4.2 and 4.3 present experimental studies on MMR and MMR-HIL, respectively.

### 4.1 Dataset and Experimental Setup

MMR and MMR-HIL are tested on a simplified regional power system in China, named CEPRI-TAS [49]. CEPRI-TAS contains 15 generators and 85 buses, with its wiring diagram shown in Fig. 3. We employ Monte-Carlo method to generate samples with faults located on various buses or transmission lines and obtain two datasets with different kinds of contingencies. The contingencies in the first dataset include N-1 and N-2 three-phase faults, while the second dataset includes N-3 three-phase fault, DC restart failure and DC bi-polar block fault. All contingencies are cleared within 150ms. The size of the two balance datasets is 4300 and 3100 respectively, and the datasets are split into training and testing sets at the ratio of 3:1. The two training datasets are contaminated by different types of FLI attacks (Sym-FLI and Asym-FLI) with injection ratios of 10%, 20% and 30%.

The encoder and decoder are implemented as convolutional neural networks. The encoder consists of three 2D convolutional layers with kernel sizes of 5×5, 5×5 and 4×4, respectively. The corresponding channels are set as 8, 16 and 32. The decoder comprises three 2D transposed convolutional layers whose kernel sizes are 5×5, 5×5 and 4×4. The step size of the convolutional and transposed convolutional layers is 2. The classifier is a multi-layer perceptron with a structure of $Z_e$-16-2, where the embedding feature dimension $Z_e$ is 64. The correction coefficient $\kappa$ is 0.03. The two balance coefficients $\alpha_1$ and $\alpha_2$ are both set as 1.0. Fuzzifier is set as 2.0. For the hyper-parameters in MMR-HIL, the annotation rate $\rho$ and the annotation frequency $T$ are 0.55% and 3. Adam is used as the optimizer and the learning rate is 0.001 for all the experiments.

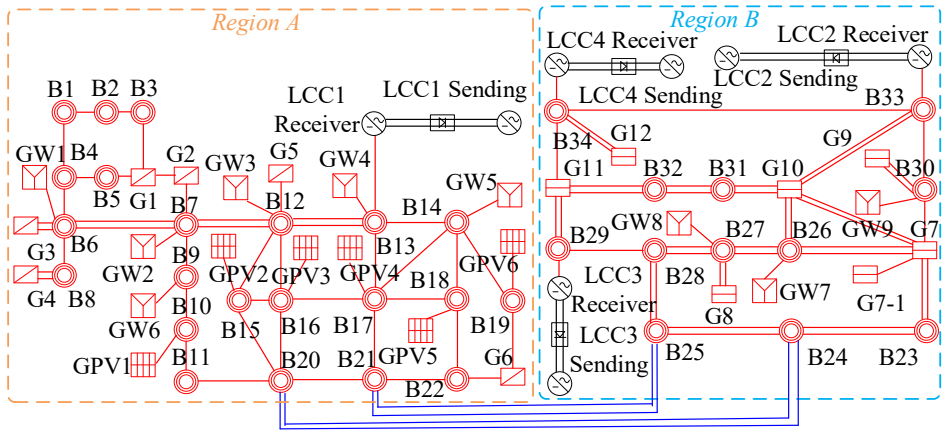

Fig. 3 The wiring diagram of CEPRI-TAS power system.

## 4.2 Experimental Studies on MMR

This section presents experimental studies conducted on MMR. Section 4.2.1 evaluates the performance of MMR through comparative experiments with other deep TSA methods. In section 4.2.2, ablation studies are conducted to analyze the contribution of each component in MMR. Parameter sensitivity analysis is performed in Section 4.2.3, while Section 4.2.4 focuses on false label correction analysis.

### 4.2.1 Model Performance

To evaluate the performance of our proposed robust TSA method, we conduct a series of comparison experiments under different FLI attacks, comparing MMR and other deep TSA methods. The comparative methods include conventional neural networks (FCN, CNN and GRU) and several popular robust learning methods, including Co-T [50], Dual-T [21], O2U [51], and D-Mix [26]. The detailed information of these methods are provided in Appendix B

**Table 1**
Accuracy performance of different deep TSA methods under N-1 and N-2 three-phase faults.

|   | Sym-FLI | | | Asym-FLI | | |
|---|---|---|---|---|---|---|
|   | 10% | 20% | 30% | 10% | 20% | 30% |
| FCN | 92.76 | 90.90 | 87.15 | 91.71 | 88.77 | 85.79 |
| CNN | 93.38 | 91.19 | 86.02 | 92.24 | 90.19 | 86.41 |
| GRU | 93.41 | 91.30 | 88.82 | 93.25 | 91.36 | 84.53 |
| Co-T | 96.10 | 95.33 | 93.36 | 95.17 | 94.79 | 93.74 |
| Dual-T | 94.68 | 91.37 | 89.72 | 92.22 | 89.46 | 86.21 |
| O2U | 95.21 | 92.53 | 90.32 | 91.17 | 90.44 | 86.28 |
| D-Mix | 97.34 | 96.10 | 93.53 | 97.01 | 95.53 | 91.79 |
| MMR | **98.62** | **97.87** | **96.73** | **98.02** | **97.35** | **95.32** |

**Table 2**
Accuracy performance of different deep TSA methods under N-3 Three-phase Fault, DC Restart Failure and DC Bi-polar Block Fault.

|   | Sym-FLI | | | Asym-FLI | | |
|---|---|---|---|---|---|---|
|   | 10% | 20% | 30% | 10% | 20% | 30% |
| FCN | 92.35 | 91.73 | 83.98 | 92.20 | 87.46 | 85.03 |
| CNN | 94.68 | 91.62 | 87.01 | 93.52 | 91.95 | 87.03 |
| GRU | 93.92 | 91.28 | 86.94 | 93.54 | 90.26 | 89.36 |
| Co-T | 96.66 | 94.74 | 92.26 | 94.03 | 93.93 | 93.16 |
| Dual-T | 94.19 | 91.26 | 86.20 | 93.15 | 91.12 | 89.70 |
| O2U | 95.27 | 93.34 | 89.42 | 93.43 | 92.67 | 88.39 |
| D-Mix | 97.04 | 96.77 | 91.73 | 97.49 | 94.86 | 90.95 |
| MMR | **98.67** | **97.21** | **96.18** | **98.00** | **97.35** | **96.53** |

Performance comparisons under different FLI attacks on the two datasets are reported in Table 1 and 2. Firstly, under the same false label injection, we can find that the accuracy of the robust learning methods (from Co-T to MMR) is significantly superior to the conventional neural networks (from FCN to GRU). This indicates that false labels can lead to severe overfitting. Without intervening in these false labels, deep TSA models will have poor performance. Among all the comparative methods, MMR demonstrates the strongest robustness against FLI. For example, in Table 1, under 30% Sym-FLI, MMR achieves assessment accuracy increments of 7.91% and 3.20% compared to the highest performance of conventional neural networks (GRU 88.82%) and robust learning methods (D-Mix 93.53%), respectively. These performance improvements suggest that utilizing unsupervised learning to rectify the misguided supervised learning not only effectively alleviates the overfitting problem caused by FLI but also demonstrates advantages compared to other robust learning methods, such as false sample detection and robust architecture. Furthermore, as the injection ratios increase, the performance of all methods tends to decrease, including MMR. However, our proposed method exhibits the smallest performance degradation. In Table 1, for Sym-FLI, as the injection ratio increases from 10% to 30%, MMR experiences only a 1.89% decrease. For comparison, the top-performing methods in conventional neural networks (GRU) and robust learning methods (Co-T) exhibit the least performance degradation for 4.39% and 2.74%. The results in Table 2 also show a similar phenomenon, suggesting that MMR has a good applicability across different types of faults. Since there is no inherent difference between the two datasets, the subsequent experiments in Section 4.2 will only be conducted on the N-1 and N-2 three-phase faults dataset.

### 4.2.2 Ablation Study

In this section, ablation studies are conducted to analyze the contribution of different components in MMR under 20% Sym-FLI and Asym-FLI. From the perspective of neural networks, MMR consists of three components: autoencoder (combination of encoder and decoder), classifier and clustering layer. Without autoencoder (AE), the input of the classifier and clustering layer will be the original transient response trajectories, which possess a very high dimensionality. If the classifier is absent in MMR, the model will degenerate into a deep clustering model [4]. Without the clustering layer, the clustering assignments cannot be obtained, resulting in the offline of the training label corrector. The ablation study results are presented in Table 3. In the following analyses, we will use the row index as the experiment index. The contrast between the 3-rd, 5-th and 2-nd, 6-th experiments shows that with the help of AE, the performance of the classifier and clustering layer has improved by 3.34%, 6.18% (3-rd, 5-th) and 17.27% (2-nd, 6-th) under Sym-FLI and Asym-FLI. These performance improvements demonstrate that the embedding features extracted by the encoder can alleviate the negative effect brought by the Curse of Dimensionality. This phenomenon can also be observed in the 7-th and 8-th experiments. Besides, since the clustering assignments predicated by the clustering layer can correct the contaminated training labels, they improve the performance of MMR by 5.79% under 20% Sym-FLI, as demonstrated in the 5-th and 8-th experiments. The comparison between the 6-th and 8-th experiments confirms that because of its unsupervised nature, the clustering module suffers from weak representation learning capability. When all three neural networks (AE, classifier and clustering layer) are employed, MMR achieves the best performance.

Table 3

Accuracy performance of MMR with different configurations under 20% Sym-FLI and Asym-FLI.

| Index | AE | Classifier | Clustering | 20% Sym-FLI | 20% Asym-FLI |
|---|---|---|---|---|---|
| 1 | × | × | × | - | - |
| 2 | × | × | ✓ | 69.76 | 69.76 |
| 3 | × | ✓ | × | 88.74 | 85.76 |
| 4 | ✓ | × | × | - | - |
| 5 | ✓ | ✓ | × | 92.08 | 91.94 |
| 6 | ✓ | × | ✓ | 87.03 | 87.03 |
| 7 | × | ✓ | ✓ | 83.58 | 81.31 |
| 8 | ✓ | ✓ | ✓ | **97.87** | **97.35** |

The reconstruction loss is a critical component in the objective functions of both the classification and clustering module. Its main function is to constrain the distribution of samples and prevent distortion in the embedding space. To analyze the influence of the reconstruction loss $L_{Rec}$ on MMR, we exclude $L_{Rec}$ from the objective functions of the classification module and the clustering module, as reported in Table 4. The incorporation of $L_{Rec}$ into $L_{CM}$ and $L_{CluM}$ leads to accuracy improvements of 0.13% and 1.20%, demonstrating the necessity of $L_{Rec}$.

Table 4

Impact of reconstruction loss on MMR's performance under 20% Sym-FLI and Asym-FLI.

| Index | $L_{Rec}$ in $L_{CM}$ | $L_{Rec}$ in $L_{CluM}$ | 20% Sym-FLI | 20% Asym-FLI |
|---|---|---|---|---|
| 1 | × | × | 94.01 | 93.75 |
| 2 | × | ✓ | 94.14 | 94.02 |
| 3 | ✓ | × | 95.21 | 95.22 |
| 4 | ✓ | ✓ | 97.87 | 97.35 |

### 4.2.3 Parameter Sensitivity Analysis

Sensitivity of important hyper-parameters in MMR has also been studied, including the two balance coefficients $\alpha_1$ and $\alpha_2$ and the correction coefficient $\kappa$. In Fig. 4, we present the performance of different settings of $\alpha_1$ and $\alpha_2$ under 20% Sym-FLI and Asym-FLI. It can be seen that when the two balance coefficients are smaller than $10^1$, MMR exhibits insensitivity to $\alpha_1$ and $\alpha_2$. However, when $\alpha_1$ and $\alpha_2$ are larger than $10^1$, the accuracy decreases rapidly. This occurs because when the balance coefficients are excessively large, the contribution of $L_{Rec}$ becomes negligible. Without the constraint of the reconstruction loss, MMR once again encounters the dilemma of embedding space distortion, resulting in severe overfitting.

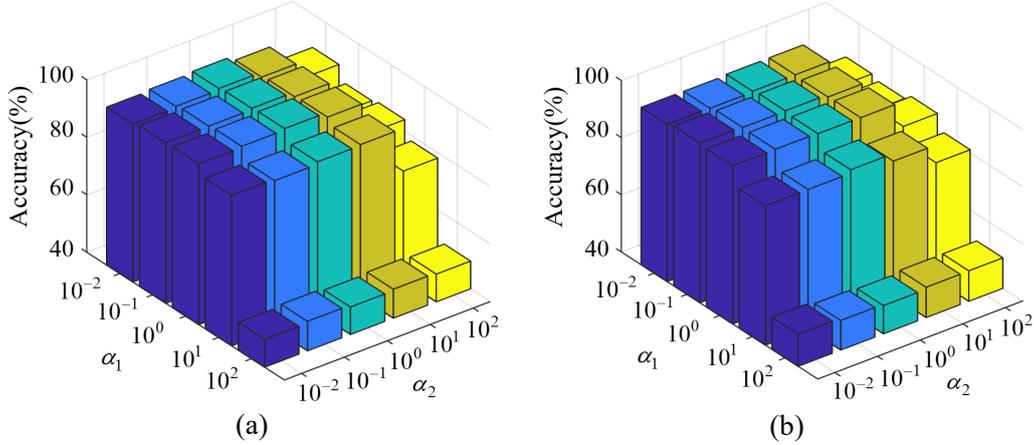

**Fig. 4** Sensitivity of $\alpha_1$ and $\alpha_2$ under (a) 20% Sym-FLI and (b) 20% Asym-FLI.

What's more, the correction coefficient $\kappa$ controls the mixture ratio of the classification predictions and the clustering assignments to the training labels, playing a crucial role in the training label correction process. Fig. 5 illustrates the variation in accuracy performance with respect to different correction coefficients $\kappa$. It can be observed that as the correction coefficient increases, the accuracy first exhibits slight fluctuations before decreasing. These results indicate that too large $\kappa$ results in overly aggressive label correction, particularly during the early stages of training when the performance of the classifier and clustering layer is poor, exacerbating overfitting. Therefore, the selection of $\kappa$ is supposed to be conservative.

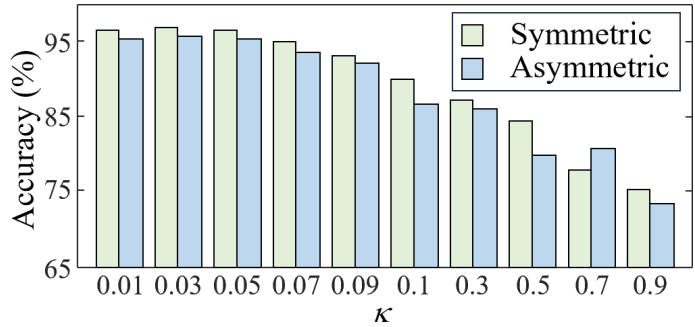

**Fig. 5** The variation of accuracy with different correction coefficients $\kappa$.

### 4.2.4 False Label Correction Analysis

In this section, we analyze the false label correction process of MMR. MMR corrects the injected false labels by integrating the classification predictions and the clustering assignments thereby improving robustness against FLI. Therefore, the proportion of false labels corrected in the training data is an important indicator of the effectiveness of MMR, reflecting its cyber resilience against FLI. The correction rate of false labels during different training stages under 30% Sym-FLI and Asym-FLI is reported in Fig. 6. We also report the correction rate under other FLI attacks in Appendix C. It can be seen that as the training progresses, the correction rate gradually increases, reaching a peak of 97.16% and 93.42% under Sym-FLI and Asym-FLI. The experiment results indicate the effectiveness of the proposed training label correction mechanism. In other words, when FLI occurs, over 90% of the contaminated labels can be restored. What's more, Table 5 presents the correction rate of false labels at the end of training under different FLI attacks. $S_F$ and $U_F$ denote the false stable and unstable labels, while $S_T$ and $U_T$ correspond to the true labels. $S_F \rightarrow U_T$ means the false stable labels being corrected as unstable, vice versa. These results fully demonstrate the superior resilience performance of MMR. However, it is evident that some stubborn false labels resist correction, validating the necessity of the human-in-the-loop training strategy MMR-HIL we proposed.

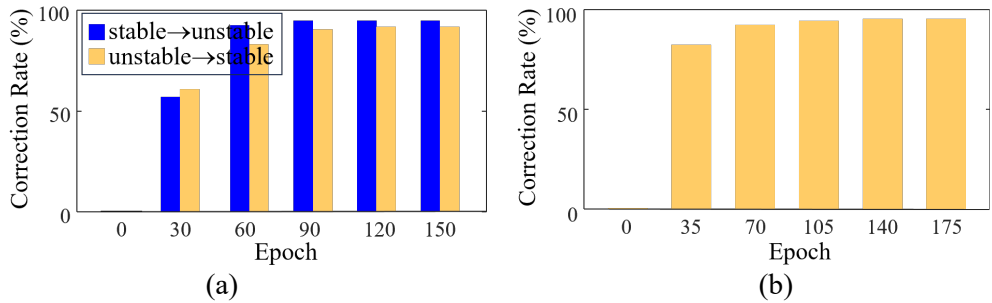

**Fig. 6** Correction rate of false labels during different training stages. (a) 30% Sym-FLI, (b) 30% Asym-FLI.

**Table 5**
False label correction rate (Cyber Resilience) of MMR under different FLI attacks.

|  | Sym-FLI | | Asym-FLI |
| --- | --- | --- | --- |
|  | $S_F \rightarrow U_T$ | $U_F \rightarrow S_T$ | $S_F \rightarrow U_T$ |
| 10% | 98.50 | 98.67 | 97.43 |
| 20% | 96.01 | 98.14 | 96.37 |
| 30% | 97.16 | 93.42 | 95.68 |

Additionally, the distribution of embedding features during different training stages under

30% Sym-FLI is visualized using *t*-SNE in Fig. 7. In the early stages of training, the distribution of training data in the embedding space appears relatively chaotic due to the presence of injected false labels. However, when ignoring the label information, it becomes apparent that there are a number of samples cluster together, as shown in the black rectangle in Fig. 7(a). These samples contribute to creating favorable conditions for the clustering layer. As training progresses, the consistency between the classification predictions and the clustering assignments improves, and some of the false labels have been corrected by the training label corrector, as indicated by the green dots in Fig. 7(b-d). At the end of the training (Fig. 7(e)), we can find that the boundary between the stable samples and unstable samples becomes distinct. Except for a few stubborn samples, the majority have been accurately classified. The visualization on the testing data, as depicted in Fig. 7(f)-(h), also demonstrates the gradual improvement in the robustness of MMR.

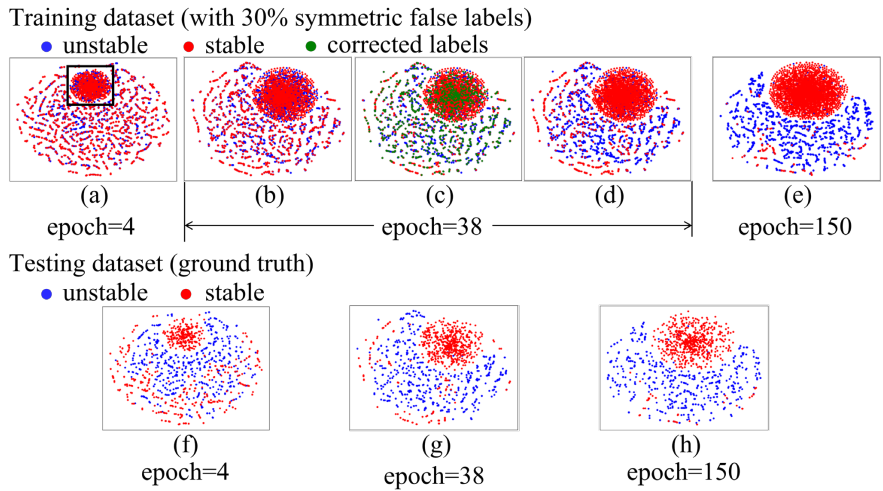

**Fig. 7** Visualization of embedding feature distribution during different training stages under 30% Sym-FLI.

### 4.3 Experimental Studies on MMR-HIL

This section presents experimental studies on the proposed human-in-the-loop training strategy, MMR-HIL, including design rationality analysis (Section 4.3.1), parameter sensitivity analysis (Section 4.3.2), and false label correction analysis (Section 4.3.3).

**4.3.1 Design Rationality Analysis**

MMR-HIL involves three steps: false sample detection, re-labeling and penalized reweight optimization. Penalized reweight optimization, as a mature technique, has received thoroughly validations in other deep TSA researches [52, 53]. Therefore, we conduct a series of experiments to validate: (1) the necessity of introducing the human-in-the-loop training strategy, (2) the effectiveness of the Small-Loss assumption during false sample detection, and (3) the rationality behind using the bi-directional annotator during re-labeling.

**Table 6**
Accuracy and convergence epoch of MMR and MMR-HIL under N-1 and N-2 Three-phase Faults and the FLI-free situation.

|  | index |  | Sym-FLI | | | Asym-FLI | | |
| --- | --- | --- | --- | --- | --- | --- | --- | --- |
|  |  |  | 10% | 20% | 30% | 10% | 20% | 30% |
| Accuracy | 1 | MMR | 98.81 | 97.82 | 96.37 | 98.43 | 97.81 | 96.73 |
|  | 2 | MMR-HIL | 99.47 | 98.96 | 98.68 | 99.16 | 98.57 | 98.00 |
|  | 3 | FLI-free | | | 99.21 | | | |
| Accuracy Increment $\Delta$ | 4 | $\Delta_{MMR/free}$ | -0.40 | -1.39 | -2.84 | -0.78 | -1.40 | -2.48 |
|  | 5 | $\Delta_{HIL/MMR}$ | +0.66 | +1.14 | +2.31 | +0.73 | +0.76 | +1.27 |
|  | 6 | $\Delta_{HIL/free}$ | +0.26 | -0.25 | -0.53 | -0.05 | -0.64 | -1.21 |
| Convergence Epoch | 7 | MMR | 108 | 126 | 163 | 133 | 145 | 189 |
|  | 8 | MMR-HIL | 52 | 57 | 68 | 71 | 78 | 93 |
|  | 9 | FLI-free | | | 52 | | | |
| Convergence Epoch Increment $k$ | 10 | $k_{MMR/free}$ | +56 | +74 | +111 | +81 | +93 | +137 |
|  | 11 | $k_{HIL/MMR}$ | -56 | -69 | -95 | -62 | -67 | -96 |
|  | 12 | $k_{HIL/free}$ | 0 | +5 | +16 | +19 | +26 | +41 |

**Table 7**
Accuracy and convergence epoch of MMR and MMR-HIL under N-3 Three-phase Fault, DC Restart Failure and DC Bi-polar Block Fault and the FLI-free situation.

|  | index |  | Sym-FLI | | | Asym-FLI | | |
| --- | --- | --- | --- | --- | --- | --- | --- | --- |
|  |  |  | 10% | 20% | 30% | 10% | 20% | 30% |
| Accuracy | 1 | MMR | 98.67 | 97.21 | 96.18 | 98.00 | 97.35 | 96.53 |
|  | 2 | MMR-HIL | 99.31 | 99.02 | 98.73 | 99.22 | 98.84 | 98.52 |
|  | 3 | FLI-free | | | 99.47 | | | |
| Accuracy Increment $\Delta$ | 4 | $\Delta_{MMR/free}$ | -0.80 | -2.26 | -3.29 | -1.47 | -2.12 | -2.94 |
|  | 5 | $\Delta_{HIL/MMR}$ | +0.64 | +1.81 | +2.55 | +1.22 | +1.49 | +1.99 |
|  | 6 | $\Delta_{HIL/free}$ | -0.16 | -0.45 | -0.74 | -0.25 | -0.63 | -0.95 |
| Convergence Epoch | 7 | MMR | 89 | 101 | 134 | 94 | 112 | 135 |
|  | 8 | MMR-HIL | 52 | 59 | 83 | 51 | 62 | 68 |
|  | 9 | FLI-free | | | 41 | | | |
| Convergence Epoch Increment $k$ | 10 | $k_{MMR/free}$ | +48 | +60 | +93 | +53 | +71 | +94 |
|  | 11 | $k_{HIL/MMR}$ | -37 | -42 | -51 | -43 | -50 | -67 |
|  | 12 | $k_{HIL/free}$ | +11 | +18 | +42 | +10 | +21 | +27 |

Firstly, we validate *the necessity of introducing the human-in-the-loop training strategy*. Despite MMR can alleviate overfitting caused by FLI, a notable gap remains in the convergence speed and assessment accuracy between MMR and FLI-free deep TSA models (referred to as FLI-free in the subsequent discussion). As an illustration, we report the accuracy and convergence epoch of FLI-free under different FLI attacks in Table 6 and 7, as well as the performance of MMR and MMR-HIL. In the two tables, $\Delta$ denotes the accuracy increment, $\Delta_{a/b}$ is the accuracy increment between model $a$ and model $b$, where a larger $\Delta$ indicates a greater improvement in performance. Meanwhile, $k$ represents the increment in convergence epoch, and $k_{a/b}$ is the convergence epoch reduction between model $a$ and model $b$, where the smaller the value of $k$, the better model performs. Based on the experimental results in the 1-st, 3-rd, 4-th and 7-th, 9-th, 10-th rows of Table 6, it is evident that compared with the convergence epoch (52) and the accuracy (99.21%) of FLI-free, there is a significant increase in the convergence epoch of MMR under both Sym-FLI and Asym-FLI, accompanied by a noticeable decrease in the assessment accuracy. For instance, when the injection ratio is 30%, the convergence epoch of Sym-FLI and Asym-FLI increases to 163 (+111) and 189 (+137), respectively, while the accuracy decreases to 96.37% (-2.84%) and 96.73% (-2.48%). Such gap, particularly in accuracy, can pose threats to the security of power systems. Fortunately, the human-in-the-loop training strategy, MMR-HIL, can effectively narrow this gap. By comparing the 1-st, 2-nd, 5-th and 7-th, 8-th, 11-th rows, we can observe that MMR-HIL can effectively improve the robustness of MMR, particularly notable in the case of 30% Sym-FLI, where the accuracy increases by 2.31%, accompanied by a reduction of 95 epochs in convergence speed. Moreover, taking FLI-free as a baseline, MMR-HIL demonstrates less degradation in accuracy performance and even surpasses the FLI-free TSA model under 10% Sym-FLI by 0.26%. Comparing the experimental results across different injection ratios, we can observe that MMR-HIL achieves greater performance gains as the injection ratio increases than MMR. For Sym-FLI, the decline in accuracy of MMR from 10% to 30% is 2.44%, while for MMR-HIL, it is only 0.79%. These experiments confirm that the proposed MMR-HIL can effectively narrow this gap, verifying the necessity of the human-in-the-loop training strategy.

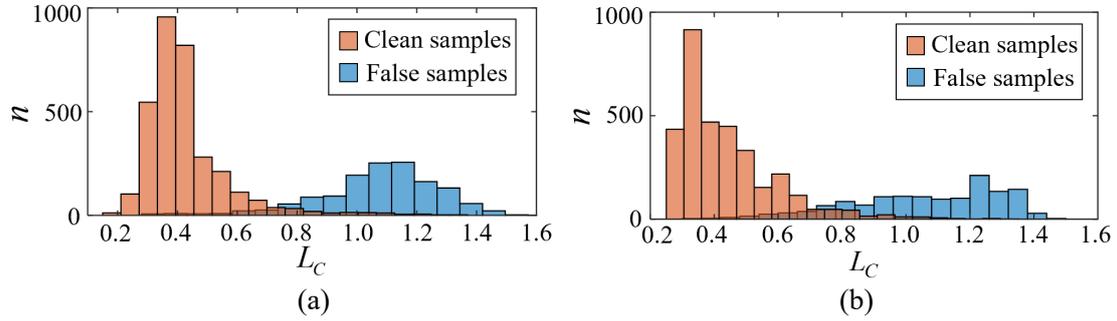

**Fig. 8** Distribution of cross-entropy loss for clean samples and false samples under different FLIs when the training epoch equals 10, (a) 30% Sym-FLI, (b) 30% Asym-FLI.

Secondly, we validate *the effectiveness of the Small-Loss assumption during false sample detection*. False sample detection plays a crucial role in MMR-HIL by distinguishing between clean and false samples. The effectiveness of the Small-Loss assumption is a necessary condition for ensuring the accurate identification of false samples. If the Small-Loss assumption is invalid, a considerable number of clean samples will be mixed with the selected false samples, severely degrading annotation efficiency and model performance. Fig. 8 illustrates the distribution of cross-entropy loss for clean and false samples under 30% Sym-FLI and Asym-FLI. The loss distributions for other FLI attacks are detailed in Appendix D. We can find that when the training epoch is 10, the loss of clean samples is noticeably smaller than that of false samples. For example, in the case of 30% Sym-FLI, the average loss of clean samples is approximately 0.37, while the average loss for false samples approaches 1.1.

Therefore, MMR-HIL can readily identify false samples by leveraging the Small-Loss assumption.

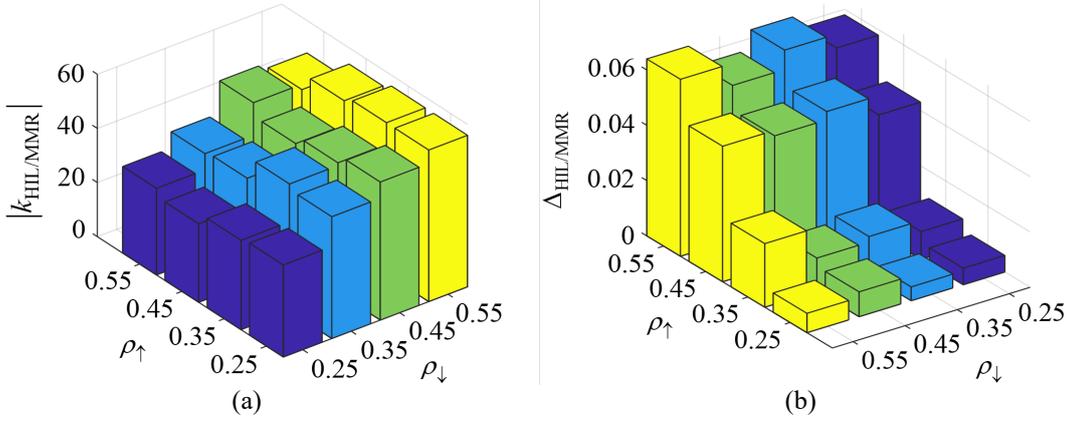

**Fig. 9** (a) The variation of the absolute convergence epoch increment $|k_{HIL/MMR}|$ with different settings of $\rho_\uparrow$ and $\rho_\downarrow$. (b) The variation of the accuracy increment $\Delta_{HIL/MMR}$ with different settings of $\rho_\uparrow$ and $\rho_\downarrow$.

Thirdly, *the rationality of the design of the bi-directional annotator* is validated. In the design of MMR-HIL, we assume that the samples selected in descending order mainly affect convergence speed, while those selected in ascending order primarily influence assessment accuracy. To verify our assumption, we have independently studied the effects of these two sampling methods. Denoting the annotation rate in descending order and ascending order as $\rho_\downarrow$ and $\rho_\uparrow$ respectively, we report the variations in accuracy increment $\Delta_{HIL/MMR}$ and the absolute value of the convergence epoch increment $|k_{HIL/MMR}|$ of MMR-HIL compared to MMR under 10% Sym-FLI with different setting of $\rho_\downarrow$ and $\rho_\uparrow$ in Fig. 9. It can be seen that increasing $\rho_\downarrow$ leads to a significant improvement in the convergence speed, while $\rho_\uparrow$ primarily influences the assessment accuracy, with a relatively minor impact on the convergence speed. The reason behind this is that most of the samples selected according to $\rho_\downarrow$ are easily identifiable false samples, which may also be corrected by the training label corrector during the MMR training process. Re-labeling these samples accelerates this correction process, primarily affecting the convergence speed. While $\rho_\uparrow$ targets on those ambiguous samples, which may also contain false samples. As a result, even after correcting the false labels of these samples, the model still requires a few epochs to capture the latent patterns. However, identifying these samples and conducting penalized reweight optimization on them can effectively refine the decision boundary, thereby improving assessment accuracy.

**4.3.2 Parameter Sensitivity Analysis**

In this section, we conduct parameter sensitivity analysis on MMR-HIL, considering both annotation rate $\rho$ and annotation frequency $T$.

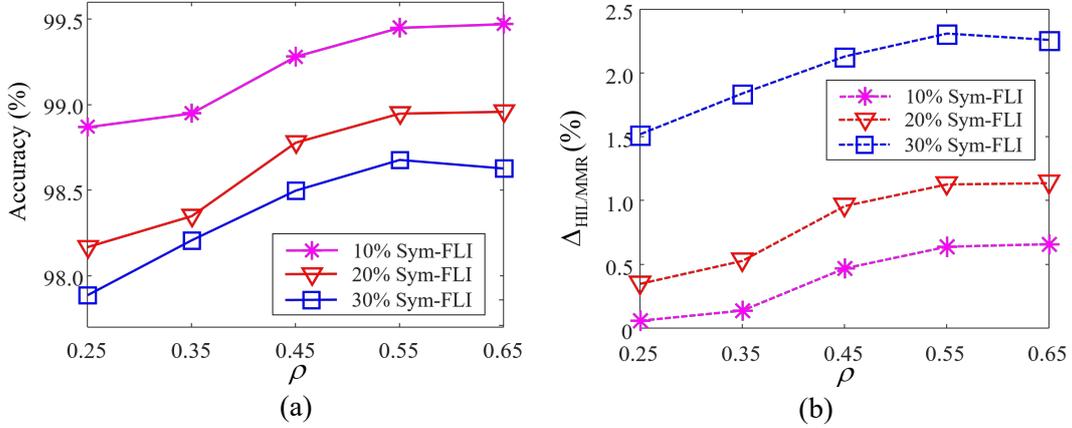

**Fig. 10** The variation of accuracy and accuracy increment under different Sym-FLI attacks with different annotation rates $\rho$. (a) Accuracy, (b) Accuracy increment.

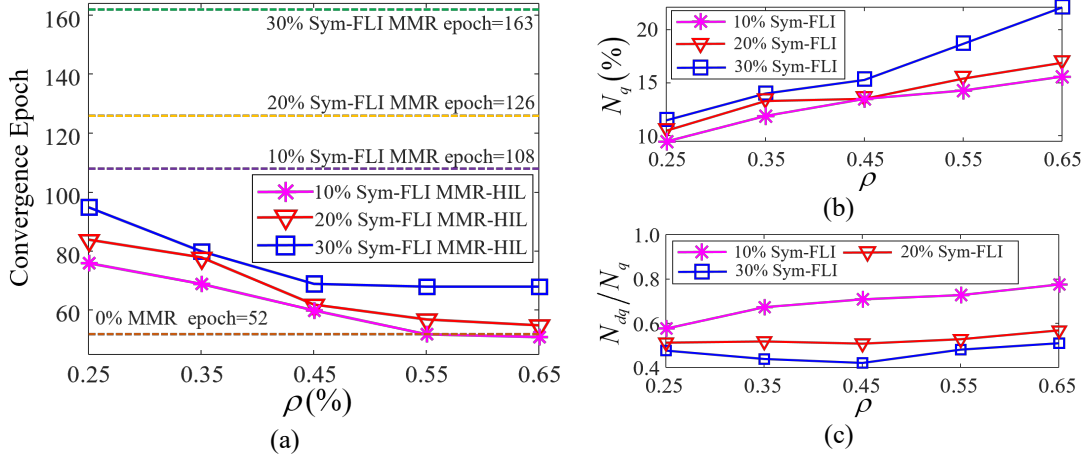

**Fig. 11** The variation of convergence epoch, proportion of query samples, and proportion of duplicate query samples in all query samples under different Sym-FLI attacks with different annotation rates $\rho$. (a) Convergence epoch, (b) proportion of query samples $N_q$, (c) proportion of duplicate query samples in all query samples $N_{dq}/N_q$.

Firstly, Fig. 10 depicts the variation of the assessment performance of MMR-HIL with different annotation rates under 10%-30% Sym-FLIs. Subfigure (a) presents the accuracy performance, while (b) shows the corresponding accuracy increment compared with MMR under the same FLI, denoted by $\Delta_{\text{HIL/MMR}}$. These experimental results indicate that, for the same level of FLI, larger annotation rate leads to larger accuracy increments. However, this increment becomes highly limited when $\rho>0.55$. Besides, across different FLI attacks, the deep TSA model benefits more from the human-in-the-loop strategy when the injection ratio is larger. This is because a higher injection ratio implies that there are more false labels in the training dataset. Consequently, compared to the case of a lower injection ratio, false samples are more easily detected and corrected. Additionally, in Fig. 11 (a), the variation of convergence epoch with the annotation rate is reported. It can be observed that similar to assessment accuracy, as the annotation rate increases, the convergence speed first increases and then stabilizes.

By analyzing Fig. 10 and 11(a), we can conclude that the performance of MMR-HIL benefits from the increase in annotation rate $\rho$. However, is a higher annotation rate always preferable? In fact, the human-in-the-loop methods [5, 54], aside from focusing on accuracy and convergence speed, also need to take into account the cost-effectiveness of annotations, i.e., achieving the maximum performance improvement with the minimum annotation cost. Based on this requirement, we report the relationship between the proportion of query samples $N_q$, the proportion of duplicate query samples $N_{dq}/N_q$ with respect to the annotation rate $\rho$ in Fig. 11(b-c). We observe that both $N_q$ and $N_{dq}/N_q$ increase as the annotation rate increases. The increase in $N_q$ implies an increase in annotation costs. Therefore, the higher the injection ratio, the higher the annotation cost required. $N_{dq}/N_q$ denotes the cost incurred due to inefficient annotation. While annotations from TSA experts can enhance the performance of deep TSA models, repetitive querying of identical samples is evidently inefficient. From this perspective, although higher injection rates result in larger $N_q$, the number of duplicate query samples is relatively low, indicating a higher efficiency.

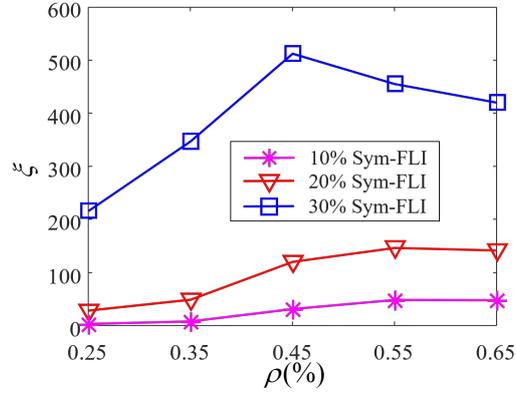

**Fig. 12** The variation of the relative efficiency under different Sym-FLI attacks with different annotation rates $\rho$.

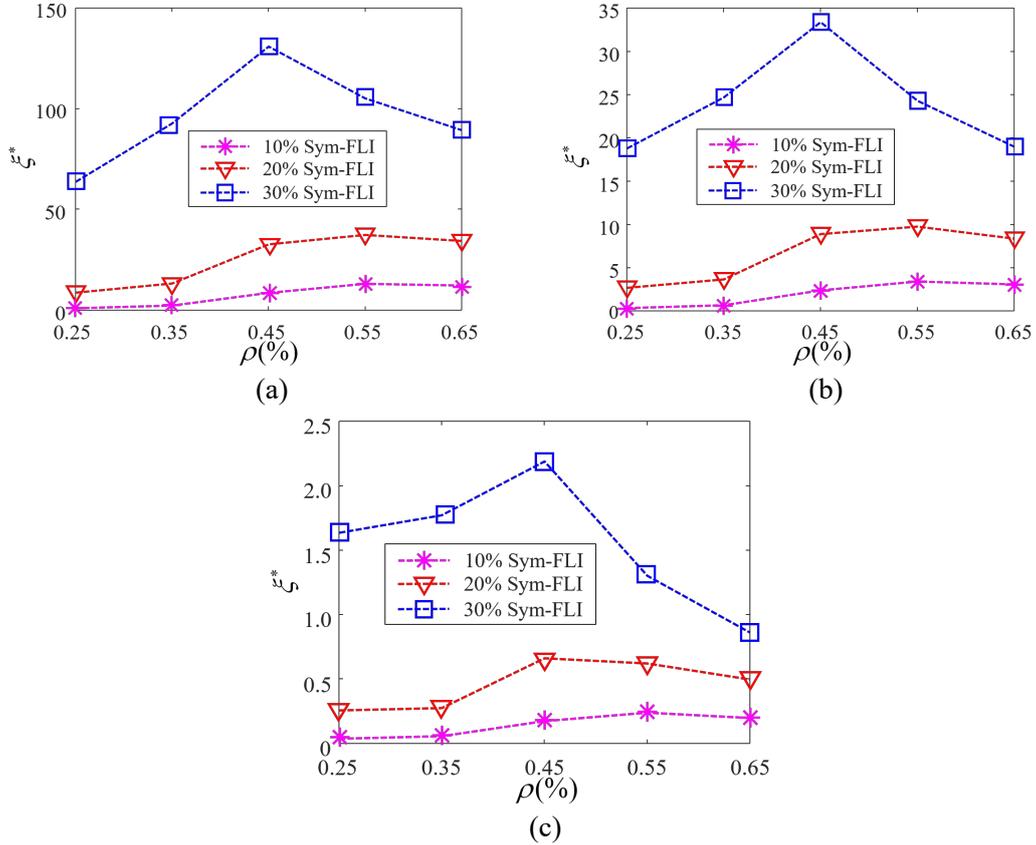

**Fig. 13** The variation of absolute efficiency of MMR-HIL under different Sym-FLI attacks with different annotation rates $\rho$. (a) $r=0.2$, (b) $r=1$, (c) $r=2$.

To quantitatively assess this trade-off relationship between performance gain and annotation cost in MMR-HIL, we introduce two evaluation metrics: relative efficiency $\xi$ and absolute efficiency $\xi^*$. By simultaneously considering the accuracy increment $\Delta_{\text{HIL/MMR}}$, the convergence epoch reduction $|k_{\text{HIL/MMR}}|$, and the proportion of duplicate query samples $N_{dq}/N_q$, relative efficiency $\xi$ is defined as:

$$\xi = \frac{\Delta_{\text{HIL/MMR}} \cdot |k_{\text{HIL/MMR}}|}{N_{dq}/N_q}. \tag{22}$$

In contrast to relative efficiency, absolute efficiency takes into account the annotation workload itself, expressed as

$$\xi^* = \frac{\xi}{N_q^r}, \tag{23}$$

where $r$ represents the degree of emphasis on the annotation workload. The higher the cost of annotation, the larger $r$ should be set. In Fig. 12 and Fig. 13, we present the variation of relative efficiency and absolute efficiency under various Sym-FLI attacks with different annotation rates. It can be observed that both types of efficiency initially increase and then decrease as the annotation rate rises. For example, in the case of 30% Sym-FLI, with an increase in the annotation rate from 0.55% to 0.65%, the accuracy only increases by 0.03%, with no change in the convergence epoch. However, these negligible performance gains come with an additional 3% of annotation cost, which is obviously highly inefficient. As a result, the relative efficiency declines from 455.3 to 420.2. The experiments about absolute efficiency can also exhibit similar phenomenon. Therefore, from the perspective of efficiency, the annotation rate should be chosen within the range of 0.45% to 0.55%. Regarding the injection ratio, higher ratios correspond to increased relative and absolute efficiency. This observation also suggests that the human-in-the-loop training strategy is well-suited for high injection ratio FLI.

What's more, to reduce annotation costs, the bi-directional annotator works every $T$ epochs. The impact of the annotation frequency $T$ on the performance of MMR-HIL under 10% Sym-

FLI is shown in Fig. 14 and Table 8. Performance degrades as $T$ increases, particularly when $T$ exceeds 5, resulting in a noticeable decline in both accuracy and convergence speed. When $T$ equals 1, the relative efficiency is very high, reaching 71.21%. However, it's worth noting that, at this point, the proportion of query samples is very large, reaching 57.4%. This indicates that over half of the samples in the training dataset need to be re-labeled, imposing a significant burden on the annotators. In consideration of the annotation workload itself, the experimental results in Table 8 reveal that the absolute efficiency at $T=1$ is only 1.32, much lower than the 3.63 achieved at $T=3$. The above analyses indicate that setting $T$ too small will cause an increase in annotation quantity, whereas setting $T$ too large will result in a degradation of model performance. Therefore, the annotation frequency $T$ is recommended to be set as 3.

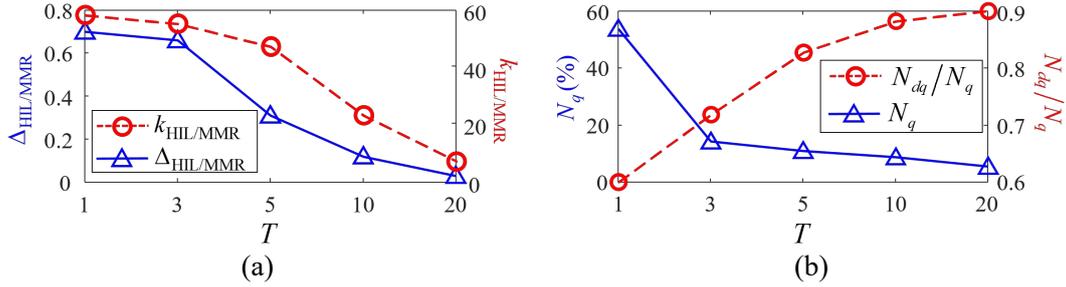

**Fig. 14** The variation of the performance gains and annotation cost with different annotation frequencies $T$. (a) performance gains, (b) annotation cost.

**Table 8**
Relative and absolute efficiency of MMR-HIL under 10% Sym-FLI with different $T$.

|  | $T=1$ | $T=3$ | $T=5$ | $T=10$ | $T=20$ |
|---|---|---|---|---|---|
| $\xi$ | **71.21** | 52.05 | 17.93 | 3.47 | 0.26 |
| $\xi^*$ | 1.32 | **3.63** | 1.63 | 0.39 | 0.05 |

### 4.3.3 False Label Correction Analysis

In this section, we analyze the resilience performance of MMR-HIL, i.e., false label correction. Table 9 reports the false label correction rate of MMR-HIL at the end of training. It can be observed that under different FLI attacks, over 98% of false labels are rectified, demonstrating the superior cyber resilience of MMR-HIL. Additionally, Table 10 presents the correction rate increment of MMR-HIL compared to MMR. The comparison results indicate that MMR-HIL exhibits stronger resilience than MMR, with a more pronounced increment as the injection ratio grows. These results once again demonstrate the effectiveness of our proposed human-in-the-loop training strategy.

**Table 9**
False label correction rate of MMR-HIL under different FLI attacks.

|     | Sym-FLI | | Asym-FLI |
| --- | --- | --- | --- |
|     | $S_F \rightarrow U_T$ | $U_F \rightarrow S_T$ | $S_F \rightarrow U_T$ |
| 10% | 99.47 | 99.61 | 99.57 |
| 20% | 99.53 | 98.33 | 98.67 |
| 30% | 99.10 | 98.16 | 98.18 |

**Table 10**
False label correction rate increment of MMR-HIL compared to MMR under different FLI attacks.

|     | Sym-FLI | | Asym-FLI |
| --- | --- | --- | --- |
|     | $S_F \rightarrow U_T$ | $U_F \rightarrow S_T$ | $S_F \rightarrow U_T$ |
| 10% | +0.97 | +0.94 | +2.14 |
| 20% | +3.52 | +0.19 | +2.30 |
| 30% | +1.94 | +4.74 | +2.50 |

## 5 Conclusion and Discussion

Focusing on false label injection in transient stability assessment, we propose a multi-module robust transient stability assessment method MMR and a human-in-the-loop training strategy MMR-HIL. MMR employs alternative training between supervised and unsupervised learning to mitigate the distortion introduced by injected false labels in the embedding feature space, leading to good performance of clustering. Integrating the clustering assignments with the classification predictions enables the correction of false labels in the training data, effectively alleviating the adverse influence of FLI on TSA performance. To further improve the accuracy and convergence speed of MMR to match that of deep TSA models trained in a FLI-free environment, we further propose a human-in-the-loop training strategy MMR-HIL. In MMR-HIL, potential false labels can be detected by modeling the classification loss with a Gaussian distribution. A bi-directional annotator is then introduced to re-label the ambiguous samples and highly likely false samples. The annotated samples are trained with a penalized reweight strategy to accelerate the training speed and improve the model assessment performance. Extensive experiments demonstrate that our proposed MMR and MMR-HIL not only exhibit robustness against FLI attacks but also effectively correct injected false labels, showcasing superior resilience.

The main scientific problem addressed by MMR is learning from false labels in multi-variate time series data, which is also a common issue in industrial settings. Therefore, both MMR and MMR-HIL hold the potential to be expanded to various fields which have FLI issues, such as electric power quality analysis and wind turbine fault diagnosis.

Although MMR and MMR-HIL effectively improve the robustness of deep TSA models against FLI, our current focus lies on balanced datasets. However, real-world TSA datasets are often imperfect. Therefore, for future research, we intend to investigate FLI issues within imbalanced and few-label TSA datasets.

## Acknowledgement

This work is supported by The National Key R&D Program of China "Response-driven intelligent enhanced analysis and control for bulk power system stability" (2021YFB2400800).